\documentclass[conference]{IEEEtran}
\IEEEoverridecommandlockouts
\usepackage{cite}
\usepackage{amsmath,amssymb,amsfonts}
\usepackage{algorithmic}
\usepackage{graphicx}
\usepackage{textcomp}
\usepackage{xcolor}
\usepackage{multirow}
\def\BibTeX{{\rm B\kern-.05em{\sc i\kern-.025em b}\kern-.08em
    T\kern-.1667em\lower.7ex\hbox{E}\kern-.125emX}}
\begin{document}

\title{Tapping the Potential of Coherence and Syntactic Features in Neural Models for Automatic Essay Scoring\\
\thanks{This paper is partially supported by Guangzhou Science and Technology Plan Project 202201010729 and GKMLP Grant No. GKLMLP-2021-001.}
}

\author{\IEEEauthorblockN{Xinying Qiu}
\IEEEauthorblockA{\textit{Guangzhou Key Laboratory of Multilingual Intelligent Processing} \\
\textit{Guangdong University of Foreign Studies}\\
Guangzhou, China \\
xy.qiu@foxmail.com}
\and
\IEEEauthorblockN{Shuxuan Liao}
\IEEEauthorblockA{\textit{School of Information Science and Technology} \\
\textit{Guangdong University of Foreign Studies}\\
Guangzhou, China \\
2399618358@qq.com}
\and
\IEEEauthorblockN{Jiajun Xie}
\IEEEauthorblockA{\textit{School of Information Science and Technology} \\
\textit{Guangdong University of Foreign Studies}\\
City, Country \\
2921109856@qq.com}
\and
\IEEEauthorblockN{Jian-Yun Nie}
\IEEEauthorblockA{\textit{Department of Computer Science and Operations Research} \\
\textit{University of Montreal}\\
Montreal, Canada \\
nie@iro.umontreal.ca}
}

\maketitle

\begin{abstract}
In the prompt-specific holistic score prediction task for Automatic Essay Scoring, the general approaches include pre-trained neural model, coherence model, and hybrid model that incorporate syntactic features with neural model. In this paper,  we propose a novel approach to extract and represent essay coherence features with prompt-learning NSP that shows to match the state-of-the-art AES coherence model, and achieves the best performance for long essays. We apply syntactic feature dense embedding to augment BERT-based model and achieve the best performance for hybrid methodology for AES. In addition, we explore various ideas to combine coherence, syntactic information and semantic embeddings, which no previous study has done before.  Our combined model also performs better than the SOTA available for combined model, even though it does not outperform our syntactic enhanced neural model. We further offer analyses that can be useful for future study.
\end{abstract}


\section{Introduction}
Automatic Essay Scoring (AES) is to automatically evaluate the quality of writing and assign grades to essays in an educational setting. The most common AES task is the prompt-specific holistic score prediction where essays written for the same prompt are evaluated with holistic scores \cite{b1, b2}. Recent research approaches towards this task include deep neural network models using RNN, CNN, or LSTM, pre-trained models, coherence models, and hybrid models that integrated multi-scale hand-crafted synatic features with neural models. These different types of methodologies present different advantages in AES tasks. Deep neural networks for AES systems using LSTM and CNN \cite{b3,b4,b5} or pretrained models such as BERT \cite{b6,b7,b8} have proved to be more robust than statistical models using hand-crafted features.  

Coherence refers to the semantic relatedness among sentences and logical order of concepts and meanings in a text \cite{b9}. When combined with neural model, coherence modeling could maintain a performance comparable to the state-of-the-art AES system \cite{b10}. In addition, incorporating coherence could improve AES system's robustness to adversarial detection (\cite{b11,b12}).

Hybrid models incorporate hand-crafted syntactic features with neural models because even though neural network can extract deep semantic meanings, they may fall short on accurately representing essay features that are critical to scoring, such as spelling and grammatical errors, readability, and other syntactic features designed specifically for the task and that have proven effective \cite{b13,b12,b14}. 

The purpose of our research is to explore the potential of BERT-based models enhanced with coherence and/or syntatic features in prompt-specific holistic scoring task for AES. In particular, we study the application of prompt-learning of next-sentence-prediction methodology \cite{b15} to model essay coherence, and the syntactic dense embeddings \cite{b16} for augmenting neural AES model. In addition, no previous study has explored the combination of semantic embeddings with coherence, and syntactic feature modeling for AES. We believe that it is important to understand how these different types of information can contribute to AES, and how their combination performs, in addition to achieving SOTA performance. 

We contribute to the study of AES system in the following directions: (1) We design a novel approach of coherence modeling for AES with prompt-learning for next sentence prediction. Experimentation shows that our model matches the state-of-the-art methods in coherence approach for AES, and performs the best for long essays. (2) We apply syntatic dense embedding which has proven successful in readability assessment to AES prediction and achieve the state-of-the-art performance in hybrid neural model for AES. (3) We examine the effect of adding coherence features together with syntactic features and semantic embedding in neural AES model. We provide comparative analysis that can be usful for future research. Our codes are available upon request.

\section{Methodology}
\subsection{Coherence feature-enhanced neural model}
\begin{figure}[h]
\begin{center}
\includegraphics[width=8cm]{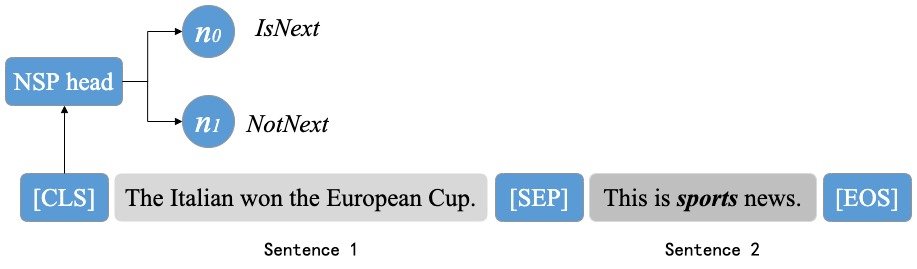}\\
\caption{\label{figNspSample} NSP task for sentence-level prompt-learning}
\end{center}
\end{figure}

NSP (Next Sentence Prediction) is a sentence-level pre-training objective in BERT that predicts whether two sentences appear consecutively within a document or not.  NSP-BERT \cite{b15} is a sentence-level prompt-learning method that aims to identify the relationship between two sentences, as illustrated in Figure~\ref{figNspSample}. 

\begin{figure}[h]
\begin{center}
\includegraphics[width=8cm]{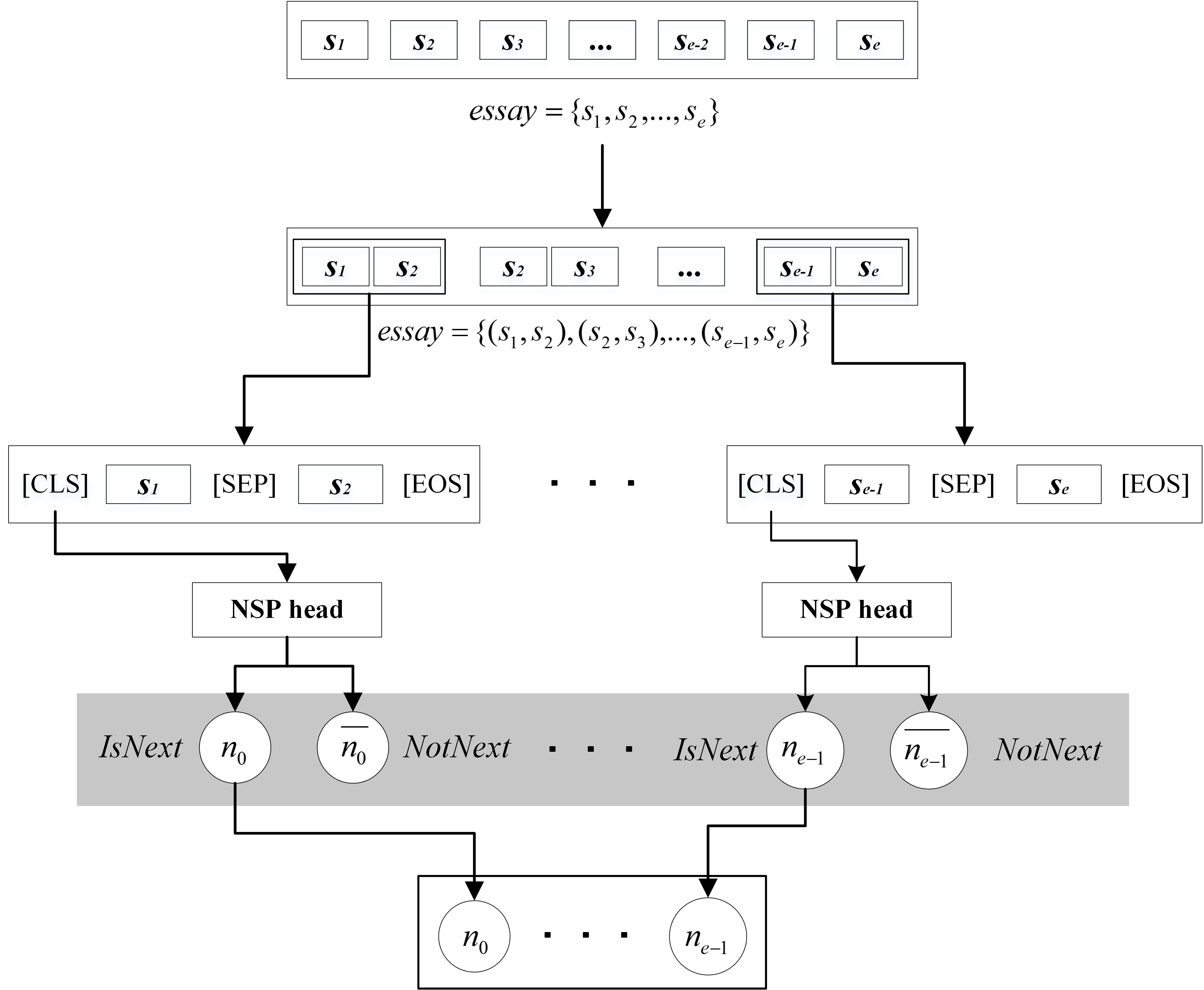}\\
\caption{\label{figNsp} Framework to extract coherence feature vector}
\end{center}
\end{figure}

\begin{figure}[h!]
\begin{center}
\includegraphics[width=6.5cm]{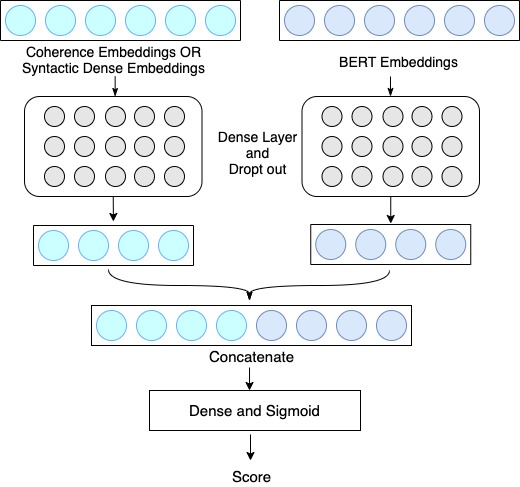}\\
\caption{\label{figDual} Dual-model to combine coherence embeddings or syntactic dense embeddings with BERT embeddings for AES predictions }
\end{center}
\end{figure}

It is empirically confirmed that NSP-BERT is not a sentence similarity matcher, but has logical reasoning ability. The model performs better when a text is formed by logically fluent sentences. In Sentence Pair Task, the NSP-BERT model has shown to outperform Zero-Shot learning and finetuning models \cite{b15}. This inspires us to use NSP-BERT model to extract the coherence feature vector from the essays using framework as illustrated in Figure~\ref{figNsp}.
We apply the NSP-BERT model to sliding windows of 2-sentence pairs and predict their ``Next Sentece'' probability. The probabilities vector is of size $e-1$ where $e$ is the number of sentences in an essay.  Since the essay lengths are different, we use the maximum length of the essays in a prompt as the uniform vector length   with zero-padding. The coherence probablity vectors are incorporated with the [CLS] output of the essay learned by finetuning BERT into a Dual-channel DNN model as described in Figure~\ref{figDual}. We name this coherenc-enhanced model as \textbf{Cohe-Enhanced BERT}.
\begin{figure}[h!]
\begin{center}
\includegraphics[width=7cm]{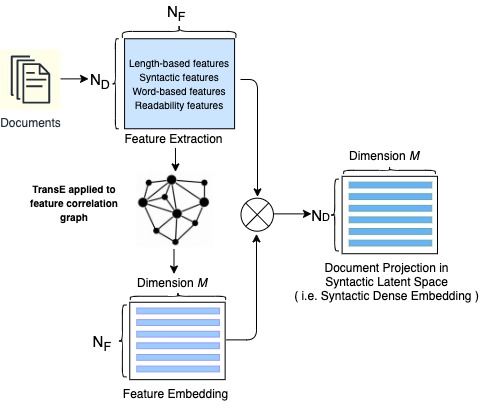}\\
\caption{\label{fig25project} Framework to construct syntactic dense embeddings with 25 AES syntactic features}
\end{center}
\end{figure}
We notice that the essay lengths vary within a single prompt. Therefore, to capture the meta features of the coherence probablity vector, we calculate 5 statistics of the probability vector without zero-padding, i.e., maximum, minimum, average, standard deviation, and perplexity which is the geometric mean. We then learn the essay embedding in the latent space of the coherence statistic features by following \cite{b16}.  

We use the essays' coherence statistic vectors to learn the correlation graph of the 5 statistics. We use gaussian binning \cite{b17} to learn the intial embedding of the 5 coherence statistical features. Then we apply TransE \cite{b18} to the correlation graph and with the initial gaussian binning embeddings to learn the optimized coherence feature embeddings.  Then with matrix multiplication, we project the coherence statistic vectors onto this latent space of the 5 probability statistics to generate essay's coherence statistical embeddings. These procedures are as illustrated in Figure~\ref{fig25project}. We then feed the embeddings into the Dual Model as in Figure~\ref{figDual}. We name this coherence-enhanced model as \textbf{Cohe\_stat-Enhanced BERT}.

\subsection{Syntatic feature-enhanced neural model}
Our methodology to enhanced neural model with handcrafted features is to apply the syntactic dense embedding method proposed by \cite{b16} which has proven successful in augmenting readability neural models. To make fair comparisons with previous work, we implement the 25 features proposed by \cite{b14} covering four categories: length-based, syntactic, word-based, and readability features. As shown in Figure~\ref{fig25project} and similar to generating coherence statisics dense embedding, we calculate the correlation graph of the 25 features, apply gaussian binning to learn the initial feature embedding, and apply TransE model to learn optimized feature embeddings. Then the essay feature vectors are projected to the latent space of the 25 features to learn syntatic dense embeddings.  The syntactic dense embeddings are fed into the Dual-channel DNN model as in Figure~\ref{figDual} to predict the grading scores of the essays. We name this model \textbf{Syntactic-Enhanced BERT}.

\subsection{Combining feature-enhanced models}
Few previous research has explored the combination of coherence modeling and syntatic features for AES. The only work closest to us is Mesgar and Strube (2018) \cite{b19} which combined their coherence vector with the feature vector from Phandi et al. (2015) \cite{b20} for neural regression prediction.  However, no semantic embeddings from the pretrained neural model are incorporated. We experiment the combination of semantic, coherence, and syntactic embeddings for AES with the following approaches:\\
\textbf{Add-Combine:} Following \cite{b8}, we add the predicted scores of BERT-based model, coherence-enhanced model, and syntatic enhanced models to a final score for evaluation.\\
\textbf{Linear-Combine:} We apply a linear activation function to the scores of different models to obtain a final score for evaluation. \\
\textbf{Concat-Combine:} We expand the Dual-channel as in Figure~\ref{figDual} to a three-channel model where multi-dense layers are applied to the BERT, coherence, and syntactic embeddings. The extracted features are then concatenated as input into dense layers for further feature extraction and score predictions.\\
\textbf{Ensemble:} Following \cite{b6}, we round the mean of the scores predicted by each of the three models (BERT-DNN, Coherence-enhanced, Syntactic-enhanced).
\section{Experiment and evaluation}
\subsection{Dataset and Experiment setting}
\label{sec:data}
\textbf{ASAP} was originally the contest data set for Automated Student Assessment Prized hosted at Kaggle in 2012. Table~\ref{data} provides the statistical description of the dataset.
We use the 5-fold cross-validation data patition presented in \cite{b3}. Following previous study \cite{b21}, for each partition, we trained each model for 100 epochs and selected the best model on the development set to apply to the test set. We reported the average QWK scores from the test sets of the 5 partitions as the model performance. This training and test procedure is repeated for each of the 8 prompts separately. Paired t-tests are used for significance tests with $p < 0.05$ to compare with BERT-DNN baseline. We use Quadratic Weighted Kappa (QWK), the official criteria of ASAP challenge, to evaluate our model.
\bgroup
\def\arraystretch{1.3} 
\begin{table}[h]
\scriptsize
\begin{center}
\begin{tabular}{c|cccc}
\hline 
\textbf{Prompt} & \textbf{\#Essays} & \textbf{Avg Length} & \textbf{Score Range} & \textbf{WordPiece Length} \\ \hline
1 & 1783 & 350 & 2-12& 649 \\
2 &1800 & 350 & 1-6 & 704  \\
3 & 1726 & 150 & 0-3 & 219 \\
4 & 1772 & 150 & 0-3 & 203 \\
5 & 1805 & 150 & 0-4 & 258 \\
6 & 1800 & 150 & 0-4 & 289 \\
7 & 1569 & 250 & 0-30 & 371 \\
8 & 723 & 650 & 0-60 & 1077 \\
\hline 
\end{tabular}
\end{center}
\caption{\label{data} Statistics of ASAP data set. }
\end{table}
\egroup

Following previous study \cite{b7}, we use two loss functions: the MSE regression loss ($L_m$) and the batch-wise ListNet ranking loss ($L_r$) as defined below, where $s_i$ and $s'_i$ are the gold score and the predicted score of the $i_{th}$ essay respectively.
\begin{flalign}
L_m & = \frac{1}{m}\sum_{i=1}^{m}(s_i - s'_i)^{2}\\
L_r & = -\sum_{j=1}^{n}P_s(j)\log(P_{s'}(j))\\
L  & = \alpha\times L_m + (1-\alpha)\times L_r
\end{flalign}

 $P_{s'}(j)$ is the top one probability defined as\\

$P_{s'}(j) = \frac{\Phi(s'_j)}{\sum_{k=1}^{n}\Phi(s'_k)}$ \\

where $\Phi$ is an exponential function as proposed by \cite{b22}. We use batch size of 32, dropout rate of 0.5 and learning rate of 0.001. We combine the two losses as in Equation (3) where $\alpha$ is tuned with the development sets.

\subsection{Baselines and Related Research}
Since our study is to improve the BERT-based neural models with coherence and syntatic features, we first implement a \textbf{BERT-DNN} baseline as shown in the right-hand channel in Figure~\ref{figDual} using the base-sized model, finetuning, and a combination loss functions as defined in Section~\ref{sec:data}. This model is similar to the BERT-DNN presented in \cite{b14} except that we use two loss functions and 100 epochs.

\textbf{(1) Coherence modeling for AES:}
 In this paper, we compare our coherence modeling methodology for AES with the following baselines:
 \begin{itemize}
\item\textbf{SKIPFLOW LSTM} \cite{b5} learns neural coherence features with SkipFlow mechanism that models relationships between snapshots of the hidden representations of LSTM network. The auxiliary coherence features are incorporated into end-to-end neural models for predition. 
\item\textbf{HierCoh} \cite{b23} is a hierarchical model consisting of sentence layers, hierarchical coherence layers, document layers and output layers. The coherence model is implemented with local attention, bilinear tensor layer, and max-coherence pooling.
\end{itemize}

\textbf{(2) Syntactic models for AES:}
We compare our syntactic feature-enhanced BERT-based model with the following hybrid models:
\begin{itemize}
\item\textbf{Qe-C-LSTM} \cite{b13} augments a hierarchical convolution recurrent neural network with different linguistic features including POS, Universal Dependency relations, Structural Well-formedness, Lexical Diversity, Sentence Cohesion, Causality, and Informativeness. The raw linguistic feature vectors go through the same neural network layers as the word embedding to be concatenated with the semantic embedding output for linear layer prediction.
\item\textbf{TSLF-ALL} \cite{b12} integrates the advantages of feature-engineered and end-to-end AES methods with a two-stage learning framework. The raw linguistic feature vectors are concatenated directly with the semantic embedding output for further prediction. Since the paper stated that the coherence score and the prompt-relevant score are removed from the framework  when applied to ASAP data set, we regard TSLF-ALL as a syntactic-feature enhanced neural AES baseline instead of a combination model with both syntactic and coherence features.
\item\textbf{BERT+Essay-level features} \cite{b14} incorporated 25 handcrafted features with BERT-based DNN-AES models where the [CLS] embedding for essay representation from BERT is concatenated with the 25 essay-level features to go through linear layer for score prediction.
\end{itemize}

\textbf{(3) Hybrid model that combines coherence and syntactic features}
There is currently only one previous research available that studies combinding coherence and syntactic features for AES. 
\textbf{EASE \& CohLSTM} \cite{b19} represents the relation between adjacent sentences by their most similar semantic states. Convolutional layer is used to extract and represent pattern of semantic changes in a text. The coherence vector is then combined with the feature vector from \textbf{EASE} \cite{b20} for neural regression prediction.  However, it does not incorporate the semantic embeddings from the pretrained neural model such as BERT.

\section{Results and Analysis}
\subsection{Coherence-enhanced model results}
We first compare the results of our coherence modeling methodology with BERT-DNN and the state-of-the-art coherence AES models as shown in Table~\ref{CoheResult}. Our methodologies are the bolded \textbf{Cohe\_stat-Enhanced BERT} and \textbf{Cohe-Enhanced BERT} in the last two rows. We can observe that both of our models significantly outperform BERT-DNN baseline with $p<0.02$ and $p<0.01$ respectively. This demonstrate that enhancing BERT-based model with our proposed coherence modeling of the essay is effective. \textbf{Cohe-Enhanced BERT} achieves an average of 0.766 which matches the state-of-the-art performance of SKIPFLOW LSTM (0.764) and HierCoh (0.763). 

Following previous study \cite{b8}, we further compare our models' performances on long essays. As shown in Table~\ref{CoheResult}, for prompts 1, 2, and 8 which have larger average essay length and WordPiece length, our coherence-enhanced models achieves an average of 0.758, better than the state-of-the-art models. This shows that our coherence feature extraction and representation methods can effectively learn the logical relations among sentence pairs within an essay. In longer essays, the logical ordering of semantics and discourse organization are more important and evident for scoring. 
\bgroup
\def\arraystretch{1.3} 
\begin{table*}[h!]
\small
\begin{center}
\begin{tabular}{c|cccccccc|c|c}
\hline \hline 
Coherence-Enhanced Model & P1 & P2 & P3 & P4 & P5 & P6 & P7 & P8 & 1-2-8 Avg &Avg QWK \\ \hline
BERT-DNN & 0.817 &	0.694	&0.679 &	0.774&	0.796& 	0.768&	0.807&	0.71&	0.74& 0.756\\
 SKIPFLOW LSTM (2018) & 0.832 & 0.684 & 0.695 & 0.788 & 0.815 & 0.810 & 0.800 & 0.697 & 0.728& 0.764 \\
HierCoh (2021) & \textbf{0.839} &	0.702	&0.711	 & 0.809	& 0.801&	0.827	& 0.82 &	0.631 & 0.724 &	0.763\\
\hline
\textbf{Cohe\_stat-Enhanced BERT} & 0.819	& 0.705 &	0.678 &	0.779 & 0.802 &	0.773 &	0.812& 0.71 &	 \textbf{0.745} & 0.76 \\
\textbf{Cohe-Enhanced BERT} & 0.83	& \textbf{0.711} &	0.689	& 0.773 &	0.797	& 0.784 &	0.812& \textbf{0.733} &	\textbf{0.758} & \textbf{0.766} \\
\hline \hline
\end{tabular}
\end{center}
\caption{\label{CoheResult} Comparing with the state-of-the-art Coherence-Enhanced Neural Models for AES.  Our models and the best performances for prompt 1, 2, and 8 and for the average are bolded. }
\end{table*}
\egroup

\bgroup
\def\arraystretch{1.3} 
\begin{table*}[h!]
\small
\begin{center}
\begin{tabular}{c|cccccccc|c|c}
\hline \hline
Syntactic-Feature-Enhanced Model & P1 & P2 & P3 & P4 & P5 & P6 & P7 & P8 & 1-2-8 Avg & Avg QWK \\ \hline
BERT-DNN & 0.817 &	0.694	&0.679 &	0.774&	0.796& 	0.768&	0.807&	0.71& 0.74 &	0.756\\
Qe-C-LSTM (2018) & 0.799&	0.631&	0.712	&0.711	 &0.801&	0.831&	0.815&	0.695& 0.708 & 0.749\footnotemark\\
TSLF-ALL (2019) & 0.852&	\textbf{0.736} &	0.731&	0.801&	0.823&	0.792&	0.762&	0.684& 0.757 &	0.773 \\
BERT + Essay-level features (2020) & 0.835 &	0.705&	0.697& 	0.776&	0.808& 	0.777&	0.818&	0.727& 0.756 & 	0.768\footnotemark \\ \hline
\textbf{Syntactic-Enhanced BERT} & \textbf{0.849} &	0.727&	0.712& 	0.779&	0.816& 	0.79&	0.833& \textbf{0.754} &	 \textbf{0.777} & \textbf{0.783} \\
\hline \hline
\end{tabular}
\end{center}
\caption{\label{featureResult} Comparing with the state-of-the-art Syntactic-enhanced Neural Models for AES. Our model and the best performances for prompt 1, 2, and 8 and for the average are bolded. }
\end{table*}
\egroup
\footnotetext[1]{Reference\cite{b13} reports an average QWK of 0.786 for their model ``Qe-C-LSTM'' in Table 4 in their paper. However, a recalculation of the average QWK from the 8 prompts gives 0.749. Therefore, we report 0.749 in our Table~\ref{featureResult}.  }
\footnotetext[2]{We could not receive responses to our inquiries from the authors of Uto et al. (2020)\cite{b14}. Therefore we reimplemented their best model following their paper as closely as we can and with their parameter setting and loss function. We report our implementation result.}

\subsection{Syntactic feature-enhanced model results}
We compare our \textbf{Syntactic-Enhanced} models with the state-of-the-art models as in Table~\ref{featureResult}. We observe that our model achieves average QWK of 0.783 which is significantly better than BERT-DNN baseline of 0.756 with $p<0.001$.

\bgroup
\def\arraystretch{1.3} 
\begin{table*}[ht]
\small
\begin{center}
\begin{tabular}{c|cccccccc|c|c}
\hline \hline
Combination Model & P1 & P2 & P3 & P4 & P5 & P6 & P7 & P8 & 1-2-8 Avg & Avg QWK \\ \hline
EASE \& CohLSTM (2018) & 0.784 & 0.654 & 0.663 & 0.788 & 0.793 & 0.794 & 0.756 & 0.646 & 0.695 & 0.728 \\
\hline
\textbf{Add-Combine} & 0.832& 	0.7 &	0.682& 	0.767&	0.810 &	0.771 &	0.807&	0.712 &0.748&0.760 \\
\textbf{Linear-Combine} & 0.827& 	0.71&	0.681& 	0.761&	0.802& 	0.776& 	0.813&	0.709& 0.749 &	0.760  \\
\textbf{Concat-Combine} & \textbf{0.847} &	\textbf{0.721} &	0.715&	0.783&	0.82&	0.79&	0.834& \textbf{0.745} & \textbf{0.771} &\textbf{0.782}\\
\textbf{Ensemble} & 0.847&	0.718&0.705	&0.789&	0.813&	0.788&	0.832&	0.741&	 0.769 & 0.779\\
\hline \hline
\end{tabular}
\end{center}
\caption{\label{combine} Comparing with the state-of-the-art combinaion model.  Our models and the best performances for prompt 1, 2, and 8 and for the average are bolded.  }
\end{table*}
\egroup

Our methodology is better than the state-of-the-art model of TSLF-ALL for overall average QWK (0.783 vs. 0.773). Similar to BERT+Essay-level model with QWK of 0.768, the TSLF-ALL model also augment neural embeddings with direct concatenation of 6 syntactic features. This shows that syntactic dense embeddings that capture the correlations between syntactic features can provide more informative knowledge to help distinguish essay quality. 

Referring to Table~\ref{featureResult} that compare performances on longer essays, we observe that our syntatic-feature enhaned model achieves better average of 0.777 for prompts 1, 2, and 8 than the state-of-the-art models. 

There may be two explanations for the advantages of our methodologies. One is that using 25 features may be more effective than 6 features and capture more discourse-level features for longer essays. Second, the syntactic dense embeddings may be a more informative representation than the raw feature vectors.

\subsection{Combining feature-enhanced models}
Currently the best combination model with both syntatic and coherence features is \textbf{EASE \& CohLSTM} \cite{b19}. We observe from Table~\ref{combine} that all of our combination methodologies out-perform \textbf{EASE \& CohLSTM} \cite{b19} with the best QWK of 0.783 compared with 0.728. We also observe that the model combination does not improve over the best  Syntactic-enhance model. One possible reason is that the coherence-enhanced model, although matching the best state-of-the-art coherence AES method, is under-performing compared with the Syntactic-enhanced model (with average QWK of 0.766 vs. 0.783). 

To investigate how useful the syntatic and coherence features are with respect to the essays' scores, we further calculate the features' Spearman correlations with the essays' true scores. Figure~\ref{cohCorre} shows that the correlations of the 5 coherence statstics of maximum, minimum, mean, standard deviation, and perplexity fall in the range of (-0.33, 0.36). 

\begin{figure}[h!]
\begin{center}
\includegraphics[width=8.5cm]{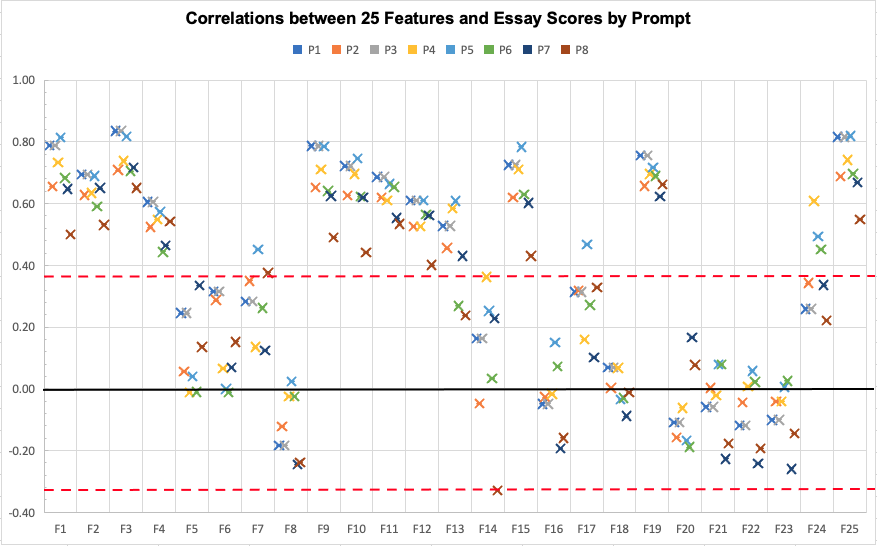}\\
\caption{\label{synCorre} Syntactic feature correlations with essay scores. The red dotted lines indicate the upper bound and lower bound of the correlations of the 5 coherence statistical features with essay scores.}
\end{center}
\end{figure}

\begin{figure}[h!]
\begin{center}
\includegraphics[width=7.5cm]{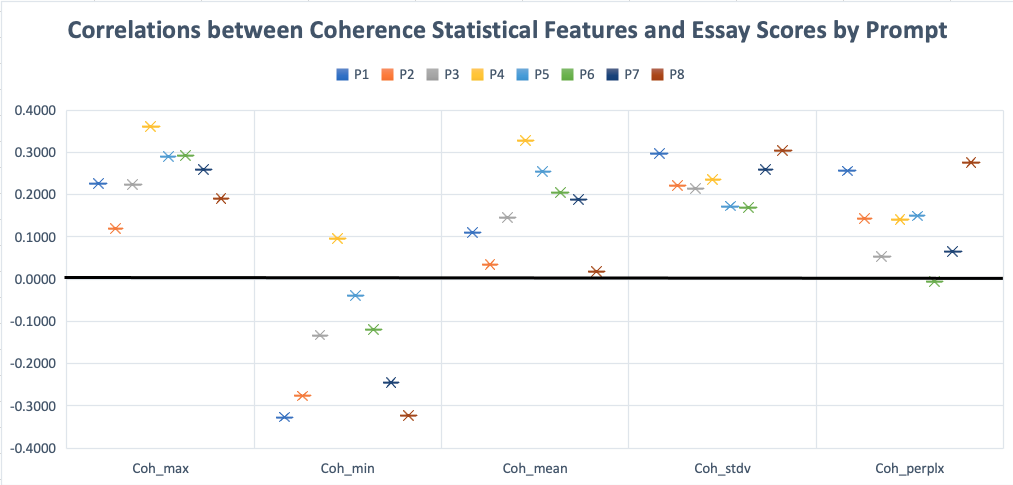}\\
\caption{\label{cohCorre} The correlations of the 5 coherence statistics with essay scores}
\end{center}
\end{figure}

In comparison,  as in Figure~\ref{synCorre}, the 25 syntactic features have correlations in the range of (-0.33, 0.84). The red dotted lines in Figure~\ref{synCorre} shows how the correlation range of coherence statistic features fits into the correlation range of the syntatic features. We observe that the syntactic features are significantly more correlated with the true scores than the coherence features. This may explain the difference in improvement between these two feature-enhanced models. Moreover, when the contribution of the coherence features in terms of relation to the true scores is not complementary to that of the syntactic features, the combination models may not add additional information than what was already captured by Syntactic-enhanced model alone.


\section{Conclusions}
In this paper, we explore the potential of coherence and syntatic features in neural models for the prompt-specific holistic scoring task of Automatic Essay Scoring.  We propose a novel approach that utilizes NSP-BERT to learn and construct coherence feature vectors. We explore a combination of coherence, syntactic, and semantic embeddings for AES which has not been studied before. We contribute to the study of AES models with the following observations:

(1) Our method to model essay coherence has shown to match the state-of-the-art performance of coherence neural model for AES and is especially effective for modeling longer essays in AES. 

(2) The application of syntactic dense embedding has shown to achieve the state-of-the-art performance for feature-enhanced hybrid models. Furthermore, the model performs especially well for  longer essays, showing that the syntactic modeling has advantage on capturing the discourse-level features more so for longer essays.

(3) We examine combining the syntactic and the coherence features and semantic embeddings in neural AES models, and outperform the state-of-the-art combined model currently available. Our combination experiments show that when syntactic features are used, adding coherence does not seem to be useful. We further analyze the different correlations with essay scores between syntactic feature and coherence features. We show that the coherence features are not as strongly correlated with the scores as the syntactic features, which may explain the strengths of syntactic-enhanced model.

Future work includes combining the strength of the pretrained model in representing shorter essay with the strength of the syntactic model for longer essay to further improve AES predictions.


\end{document}